\documentclass[final,5p,times,twocolumn]{elsarticle}

\usepackage[cmex10]{amsmath}
\usepackage{bm}
\usepackage{amsfonts}
\usepackage[ruled,vlined]{algorithm2e}
\usepackage{algcompatible}
\usepackage{subfigure}
\usepackage{subfloat}
\usepackage{graphicx}
\usepackage{xcolor}
\usepackage{multirow}
\usepackage{textcomp}
\usepackage{amssymb}
\usepackage{array}
\usepackage{txfonts}
\usepackage{bm}
\usepackage{enumerate}
\usepackage{stfloats}
\usepackage{algorithmicx,algpseudocode}
\usepackage{verbatim}
\usepackage{colortbl}
\usepackage{makecell}
\usepackage{soul}

\usepackage{setspace}
\newcommand{\PreserveBackslash}[1]{\let\temp=\\#1\let\\=\temp}

\newcolumntype{C}[1]{>{\PreserveBackslash\centering}p{#1}}
\newcolumntype{R}[1]{>{\PreserveBackslash\raggedleft}p{#1}}
\newcolumntype{L}[1]{>{\PreserveBackslash\raggedright}p{#1}}
\usepackage{xcolor}

\usepackage[pagebackref=false,breaklinks=true,letterpaper=true,bookmarks=false,colorlinks=false]{hyperref}

\journal{}

\begin{document}

\begin{frontmatter}



\title{Towards the in-situ Trunk Identification and Length Measurement of \\ Sea Cucumbers via B\'{e}zier Curve Modelling}


\author[OUC2]{Shuaixin Liu}
\ead{liushuaixin@stu.ouc.edu.cn}

\author[OUC2]{Kunqian Li\corref{cor1}}
\ead{likunqian@ouc.edu.cn}

\author[OUC2]{Yilin Ding}
\ead{dingyilin@stu.ouc.edu.cn}

\author[OUC2]{Kuangwei Xu}
\ead{xukuangwei@stu.ouc.edu.cn}

\author[OUC3]{Qianli Jiang}
\ead{jiangqianli@ouc.edu.cn}

\author[WIN]{Q. M. Jonathan Wu}
\ead{jwu@uwindsor.ca}

\author[OUC2,OUC3]{Dalei Song\corref{cor1}}
\ead{songdalei@ouc.edu.cn}

\cortext[cor1]{Corresponding author}

\address[OUC2]{College of Engineering, Ocean University of China, Qingdao 266404, China.}

\address[OUC3]{Institute for Advanced Ocean Study, Ocean University of China, Qingdao 266100, China.}

\address[WIN]{Department of Electrical and Computer Engineering, University of Windsor, Windsor, ON N9B 3P4, Canada.}

\begin{abstract}

We introduce a novel vision-based framework for in-situ trunk identification and length measurement of sea cucumbers, which plays a crucial role in the monitoring of marine ranching resources and mechanized harvesting. To model sea cucumber trunk curves with varying degrees of bending, we utilize the parametric B\'{e}zier curve due to its computational simplicity, stability, and extensive range of transformation possibilities. Then, we propose an end-to-end unified framework that combines parametric B\'{e}zier curve modeling with the widely used You-Only-Look-Once (YOLO) pipeline, abbreviated as TISC-Net, and incorporates effective funnel activation and efficient multi-scale attention modules to enhance curve feature perception and learning. Furthermore, we propose incorporating trunk endpoint loss as an additional constraint to effectively mitigate the impact of endpoint deviations on the overall curve. Finally, by utilizing the depth information of pixels located along the trunk curve captured by a binocular camera, we propose accurately estimating the in-situ length of sea cucumbers through space curve integration. We established two challenging benchmark datasets for curve-based in-situ sea cucumber trunk identification. These datasets consist of over 1,000 real-world marine environment images of sea cucumbers, accompanied by B\'{e}zier format annotations. We conduct evaluation on SC-ISTI, for which our method achieves mAP50 above 0.9 on both object detection and trunk identification tasks. Extensive length measurement experiments demonstrate that the average absolute relative error is around 0.15. The new benchmarks, source code, and pre-trained models are available on the project home page: \url{https://github.com/OUCVisionGroup/TISC-Net}.

\end{abstract}

\begin{keyword}
Intelligent sea cucumber trunk identification, in-situ length measurement, B\'{e}zier curve modelling, underwater compute vision, aquaculture.
\end{keyword}

\end{frontmatter}

\section{Introduction}

\begin{figure}[t]
  \centering
  \vspace{-0.6em}
   \includegraphics[width=1\linewidth]{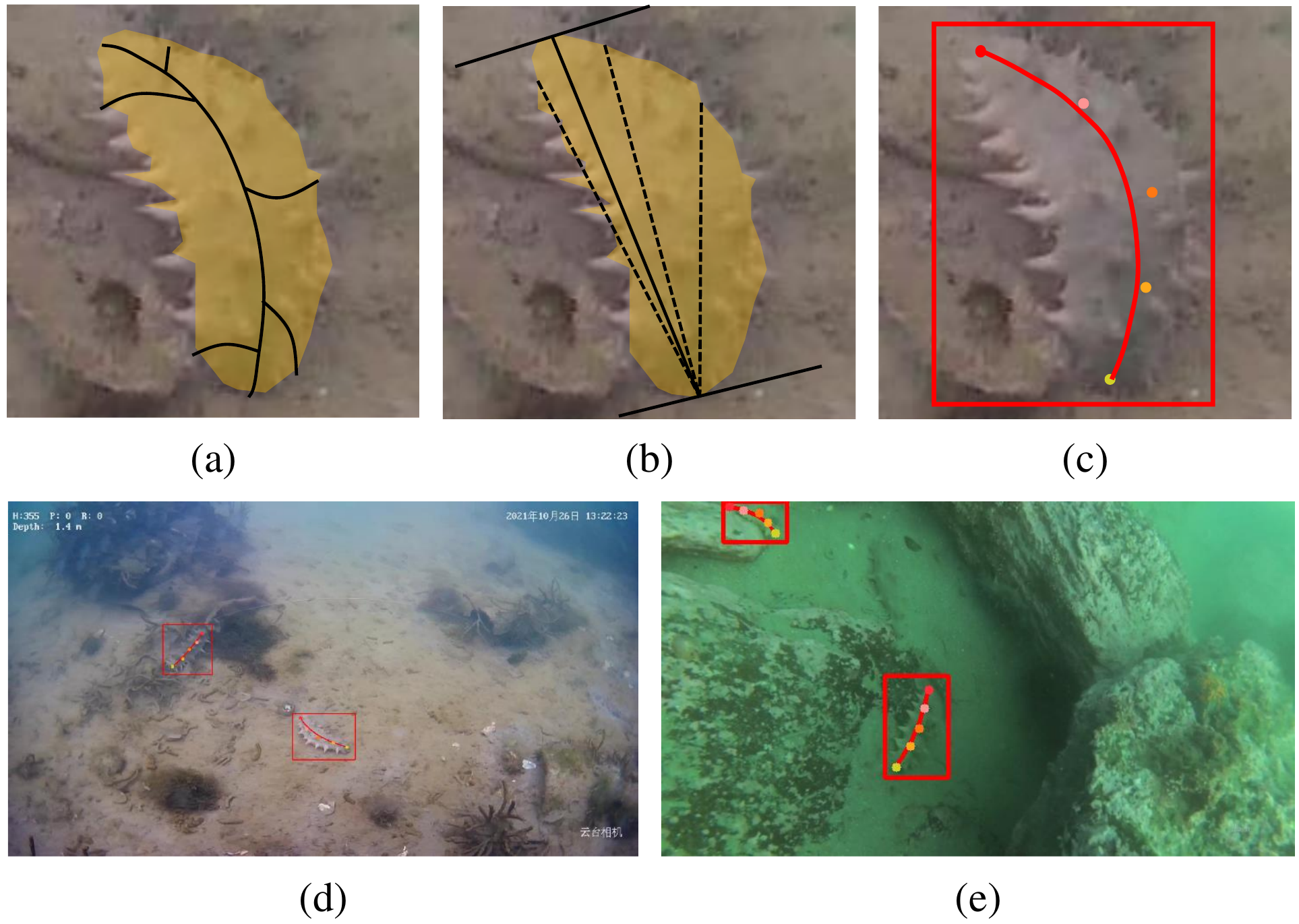}
   \caption{Traditional trunk identification strategies and the B\'{e}zier curve-based strategy that studied in this paper. (a) Morphological skeleton-based method. (b) Rotating calipers-based method. (c) The quartic B\'{e}zier curve defined by 5 control points, which can effectively fit the sea cucumber trunk shape and wrap the trunk in a bounding box. The trunk identification results of the proposed TISC-Net are exemplified in (d) and (e).}
   \label{fig:strategies}

\end{figure}

In recent years, sea cucumbers have emerged as a valuable fishery resource and play a pivotal role due to their exceptional nutritional value and robust market demand. Marine ranching, as an innovative model for the advancement of aquaculture, provides a sustainable platform for sea cucumber cultivation~\citep{zion2012use,ranjan2023effects}. The rapidly expanding field of marine ranching currently necessitates the implementation of highly efficient monitoring systems for its resources. As the examples in Fig. \ref{fig:strategies} presented, for marine ranches with sea cucumber breeding as the main industry, trunk length measurement is an important indicator to evaluate the growth and resource amount. Besides, the trunk length is also a crucial factor in assessing the maturity level during sea cucumber harvest.

The current resource survey primarily relies on manual methods. After pre-locating the station, the sampling area is manually delineated using circumference and strip methods to assess sea cucumber quantity, specifications, and growth status~\citep{cutajar2022culturing}. When manually measuring the length of sea cucumbers, the plasticity of their bodies and the stress response (resulting in the elimination of internal organs) pose challenges to obtaining accurate measurements. In addition, time out of the water significantly changes wet weight measurements~\citep{harper2020standard}, resulting in varying techniques that lack consistency. Therefore, the manual resource survey method, due to its high cost and low spatio-temporal resolution \citep{qiao2017automatic}, is not suitable for meeting the comprehensive monitoring needs of sea cucumber resources throughout the entire production cycle of marine ranching. Intelligent and unmanned technologies create new opportunities for the production of intelligent fisheries~\mbox{\citep{shi2020automatic, wu2022application}}. An underwater mobile observation platform, equipped with an intelligent vision algorithm for in-situ non-contact measurement of trunk length, is better suited for this specific task scenario.

\begin{figure}[t]
  \centering
  
   \includegraphics[width=1\linewidth]{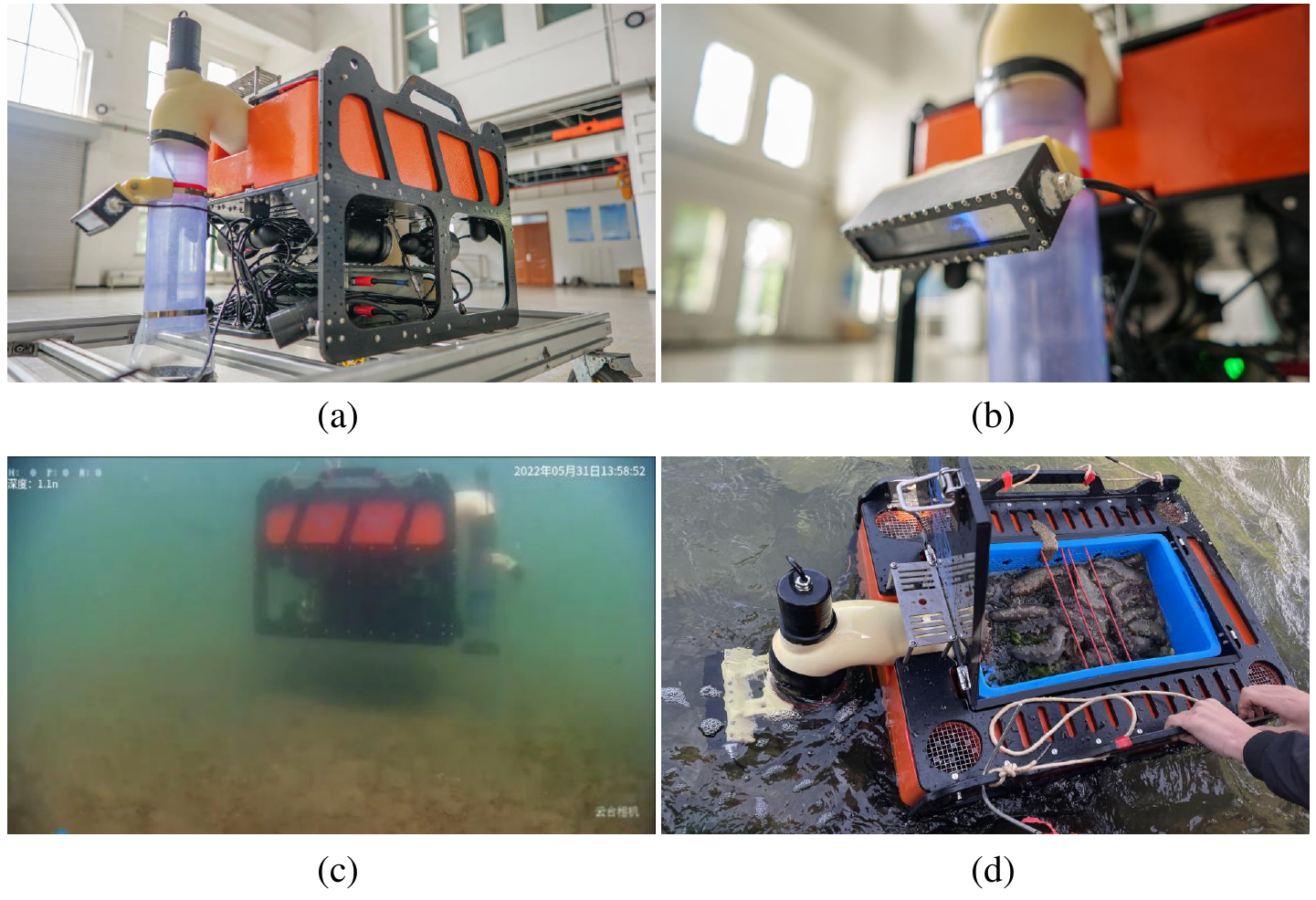}
   \caption{The sea cucumber harvesting robot developed by the Fab U+ group of Ocean University of China. (a) The overall structure of the sea cucumber harvesting robot. (b) The depth vision systems with a ZED binocular camera. (c) The robot is working in the marine ranch. (d) Recycling the sea cucumber harvesting robot and unloading the sea cucumbers.}
   \label{fig:robot}
  
\end{figure}

The utilization of underwater vehicles, including remote operated vehicles (ROVs) and autonomous underwater vehicles (AUVs), equipped with visual capabilities, has demonstrated remarkable achievements in underwater observation and operating tasks \citep{zhou2023underwater}, introducing novel strategies and perspectives to the farming of sea cucumbers in marine ranches~\citep{garcia2020automatic}. To address the safety concerns associated with manual sea cucumber harvesting underwater, we, the Fab U+ group from Ocean University of China, have developed a highly efficient robot for automated sea cucumber harvesting. The sea cucumber harvesting robot is shown in Fig. \ref{fig:robot}. As shown in Fig. \ref{fig:robot} (b), a binocular vision system was incorporated into the sea cucumber harvesting robot, enabling both visual comprehension and perception of spatial distance. This creates hardware conditions for its non-contact in-situ trunk length measurement during resource assessment and harvesting.

Unfortunately, in terms of software support, the visual perception and measurement algorithm for underwater targets is still in its early stages compared to the rapidly development in the air. The poor visual quality of underwater images and highly dynamic imaging environments are the main challenges~\citep{li2022beyond}. Thanks to their exceptional ability to represent task-oriented features, when the in-air schemes are directly applied to underwater scenarios, the deep learning-based classical approaches \citep{ren2015faster, Redmon2016YOLOv1, lyu2022rtmdet} achieve promising results in underwater object detection. Recently, specialized deep networks have been developed for fish detection~\citep{ranjan2023effects}, automatic sizing~\citep{munoz2018enhanced}, length measurement\mbox{~\citep{shi2020automatic}}, etc. 
However, an effective method to identify the trunk and measure the length of sea cucumbers is still lacking.

In Fig. \ref{fig:strategies} (a) and (b), we present two traditional and widely adopted trunk extraction strategies, i.e., the morphological skeleton-based \citep{garcia2020automatic} and the rotating caliper-based \citep{magdeev2022extraction} methods. However, both of them require segmentation masks as the prior knowledge, which mean they need to first segment out the sea cucumbers. In fact, the segmentation for underwater scenarios itself remains an unsolved problem in the field of underwater vision and requires a large amount of labor-cost annotations for model training, especially for the small and indiscernible sea cucumber targets. 

As shown in Fig. \ref{fig:strategies} (c), we observe that the classic quartic B\'{e}zier curves, with sufficient freedom degrees of parameterizing the deformations of trunk curves of sea cucumber in marine ranching, remain low computation complexity and high stability. The ease of unifying and optimizing B\'{e}zier curve modeling enables the end-to-end network to simultaneously detect targets and trunk curves. Thus, a unified deep Trunk Identification Network for Sea Cucumbers (TISC-Net) is designed to identify the trunk curves while minimizing the need for labor-intensive annotation during model training. Then, accurate in-situ length estimates can be achieved by further combining depth information from binocular cameras. Our main contributions are summarized as follows:

\begin{itemize}
\setlength{\itemsep}{-1pt} 
    \item We propose an end-to-end framework, named TISC-Net, which integrates B\'{e}zier curve modeling and a YOLO-like architecture for the joint detection and trunk identification of sea cucumbers. The TISC-Net model utilizes cost-effective annotations and effectively captures the varying degrees of curvature in trunk curves.
    
    \item Incorporating the FReLU activation function and the Efficient Multi-scale Attention (EMA) module into TISC-Net, we enhance the network's sensitivity to local features and improve its spatial context awareness for underwater visual perception tasks. Furthermore, a novel endpoint loss function is designed as auxiliary constraints to enhance the robustness of the B\'{e}zier curve.
    
    \item To promote the research, we have established two annotated benchmarks: the Sea Cucumber In-situ Trunk Identification (SC-ISTI) dataset consisting of $462$ images captured by an underwater robot in the sea cucumber habitat, and SC-DUO dataset comprising $1,023$ images derived from the Detecting Underwater Objects (DUO) dataset~\citep{DUOliu2021dataset}. Experiments and ablation studies on both benchmarks verified the good performance of our method.
    
    \item We have developed a pipeline for intelligent in-situ sea cucumber length measurement, utilizing the proposed TISC-Net, and rigorously assessed its performance in an authentic underwater environment.
\end{itemize}

\section{Related Works}\label{Related Works}
\subsection{Object Detection and the Underwater Attempts}
The objective of general object detection is to accurately localize and classify predefined objects within a given image. Since the introduction of Convolutional Neural Networks (CNNs), which have demonstrated remarkable in learning feature representations from visual data, deep CNN methods have come to dominate the field of object detection \citep{Zou202320Survey}. Recently, the increasing popularity and application of underwater visual observation has sparked a rising interest in underwater object detection techniques \citep{Xu2023review}. Considering the characteristics of underwater imaging and the application requirements of such special scenarios, researchers have carried out many researches on improving the feature representation under harsh imaging conditions  \citep{fan2020dual, chen2022swipenet}, lightweight detection \citep{yeh2021lightweight}, sample augmentation strategy \citep{lin2020roimix}, etc. The main targets of underwater object detection are fishes \citep{zion2012use, yang2021computer}, economic aquatic products \citep{fan2020dual, chen2022swipenet}, marine debris \citep{Fulton2019MarineTrash, yeh2021lightweight}, etc. However, the more complex task of underwater visual perception involving target detection as a relay or related task, such as the length measurement problem discussed in this paper, is rarely addressed. In this study, we propose a novel unified framework that seamlessly integrates object detection with trunk identification of sea cucumber as a downstream task in underwater scenes.

\subsection{Generic Object Skeleton Detection}
Object trunk identification, also widely known as object skeleton detection or extraction, is a valuable yet challenging vision task in various real-world scenarios. Early skeleton extraction methods focusing on pre-segmented images have been comprehensively studied \citep{Saha2016Survey,fernandes2020deep}. However, due to the inherent challenges associated with segmenting natural images captured in uncontrolled environments, these methods can only be effectively applied in scenarios with a relatively simple background \citep{garcia2020automatic,abinaya2022deep}. After that, traditional learning schemes with hand-crafted features were proposed \citep{Tsogkas2012learningsymmetry, Sironi2014centerline}. Unfortunately, these algorithms cannot handle complex structures, overcome background interference, or achieve satisfactory speed.

More recently, generic object skeleton extraction with deep learning framework has made great progress \citep{CMM2020Simultaneous, Liu2021Skeleton}. However, we notice that the existing methods primarily concentrate on extracting the skeletal structures of salient objects in the image \citep{CMM2020Simultaneous}, which is totally different from the sea cucumber trunk identification problem. The images in Fig. \ref{fig:strategies} illustrate that the sea cucumber exhibits a remarkable resemblance to the background, making it challenging to discern, particularly due to its relatively small scale compared to the coverage scale of the whole image. Moreover, existing generic object skeleton extraction methods devote significant effort to achieving thinner, stronger, and continuous skeletons with clear intersection points \citep{CMM2020Simultaneous}, resulting in increasingly sophisticated but intricate models. In fact, what is truly required for sea cucumber trunk identification is a unified framework that can effectively distinguish sea cucumbers from complex underwater scenarios and provide concise trunk modelling for efficient deployment.

\subsection{Specialized Object Skeleton Detection}
Specialized object skeleton detection schemes, such as human skeleton detection (also known as human pose estimation) \citep{zheng2023deep, jiang2023rtmpose}, have garnered significant attention in recent years and found applications in tasks like action recognition. For aquatic organisms, trunk identification has only been explored in the task of measuring the length of fishes \citep{white2006automated, shi2020automatic, fernandes2020deep, li2020nonintrusive}. However, the vast majority of current schemes are designed for artificial environments with a 2D imaging plane, such as conveyors \citep{white2006automated}, pipes \citep{garcia2020automatic}, studios \citep{fernandes2020deep}, tanks \citep{li2020nonintrusive}, etc. The existing 3D dimension measurement schemes based on binocular camera systems primarily focus on controlled laboratory scenes \citep{shi2020automatic, deng2022automatic} and simplistic in-situ environments \citep{hu20243d}. These approaches face challenges in adapting to complex underwater application scenarios and fail to fully consider the impact of posture changes of targets. The examples shown in Fig. \ref{fig:strategies} indicate that, for the in-situ identification of sea cucumber trunks, complex underwater backgrounds and various bends are the primary challenges.

\subsection{Curve Detection and Modelling}\label{Curve Detection and Modelling}
The curve detection and modelling, which aims to identify and extract various types of curves, contours, or boundaries from images, serves as a natural and valuable tool for describing diverse shape deformations in challenging natural environments. These curve detection techniques have been extensively explored in numerous applications, including lane detection\citep{tabelini2021polylanenet, Rethinking_Bezier_curve}, scene text detection\citep{Yuan2020Follow, Liu2022ABCNetv2}, and surface wave detection in remote sensing~\citep{dai2022study}, etc. 

Due to the exceptional representational capacity of bending transformation, curves serve as a comprehensive depiction of the sea cucumber trunk, facilitating predictions regarding diverse trunk variations. However, the unique nature of the sea cucumber trunk identification task and complicated habitat scene poses challenges in directly applying curve detection and modelling schemes commonly used in above tasks. The challenge is twofold. Firstly, as shown in Fig. \ref{fig:strategies}, the detection of the relative small sea cucumber targets in the free view of underwater vehicles poses a significant challenge for the curve modelling schemes designed for large-scale targets, such as those employed in above tasks \citep{Rethinking_Bezier_curve, Liu2022ABCNetv2, dai2022study}. Secondly, these methods do not clearly indicate the endpoints of the predicted curves, which may be deemed unnecessary or relatively less significant for tasks such as lane detection \citep{tabelini2021polylanenet, Rethinking_Bezier_curve} and surface wave detection problems \citep{dai2022study}. But obviously, the determination of the endpoint of the predicted curve plays a crucial role in ensuring accurate length measurement, which is the downstream task of curve detection and modelling. In this paper, we propose an end-to-end framework for sea cucumber trunk identification based on the parametric B\'{e}zier curves modelling. This framework employs a target-to-curve scheme and incorporates innovative endpoint constraints to effectively address the aforementioned challenges.

\begin{figure*}[t]
  \centering
   \includegraphics[width=0.9\linewidth]{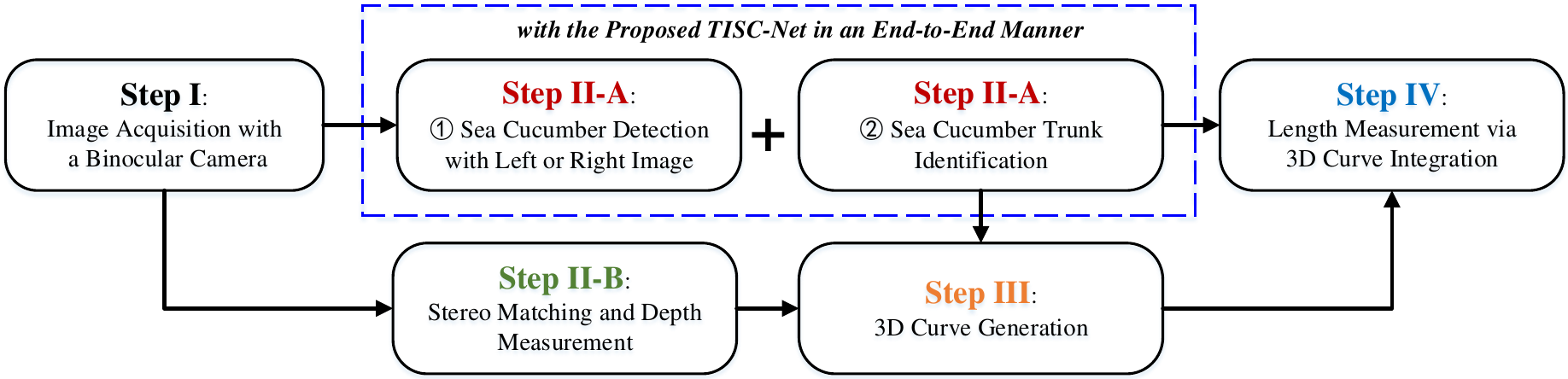}
   \caption{Pipeline of the vision-based in-situ length measurement framework with the proposed TISC-Net for end-to-end and joint sea cucumber detection and trunk identification. With stereo matching for pixel-wise depth measurement, we finally perform length measurement through integration on the 3D trunk curve, achieved by merging the 2D curve with the pixel depth.}
   \label{fig:pipeline}
\end{figure*}

\section{Proposed Method}
\subsection{Pipeline of in-situ Length Measurement of Sea Cucumbers}
In this study, we have developed a practical methodology for in-situ measurement of sea cucumber length using computer vision, as illustrated in Fig. \ref{fig:pipeline}. Utilizing a binocular camera as an in-situ underwater image collector, we propose to conduct trunk identification on the 2D plane of either the left or right image, followed by measuring the length in 3D space achieved by stereo matching for pixel-wise depth. Specifically, we propose a unified end-to-end deep learning framework based on B\'{e}zier curve modelling for simultaneous sea cucumber detection and trunk identification, thus obtaining continuous trunk pixel coordinates of the sea cucumbers in the 2D plane. Then, by merging the depth information obtained through stereo matching, we can ultimately transform the measurement of length into a 3D curve integration problem.

\begin{figure}[t]
  \centering
   \includegraphics[width=1\linewidth]{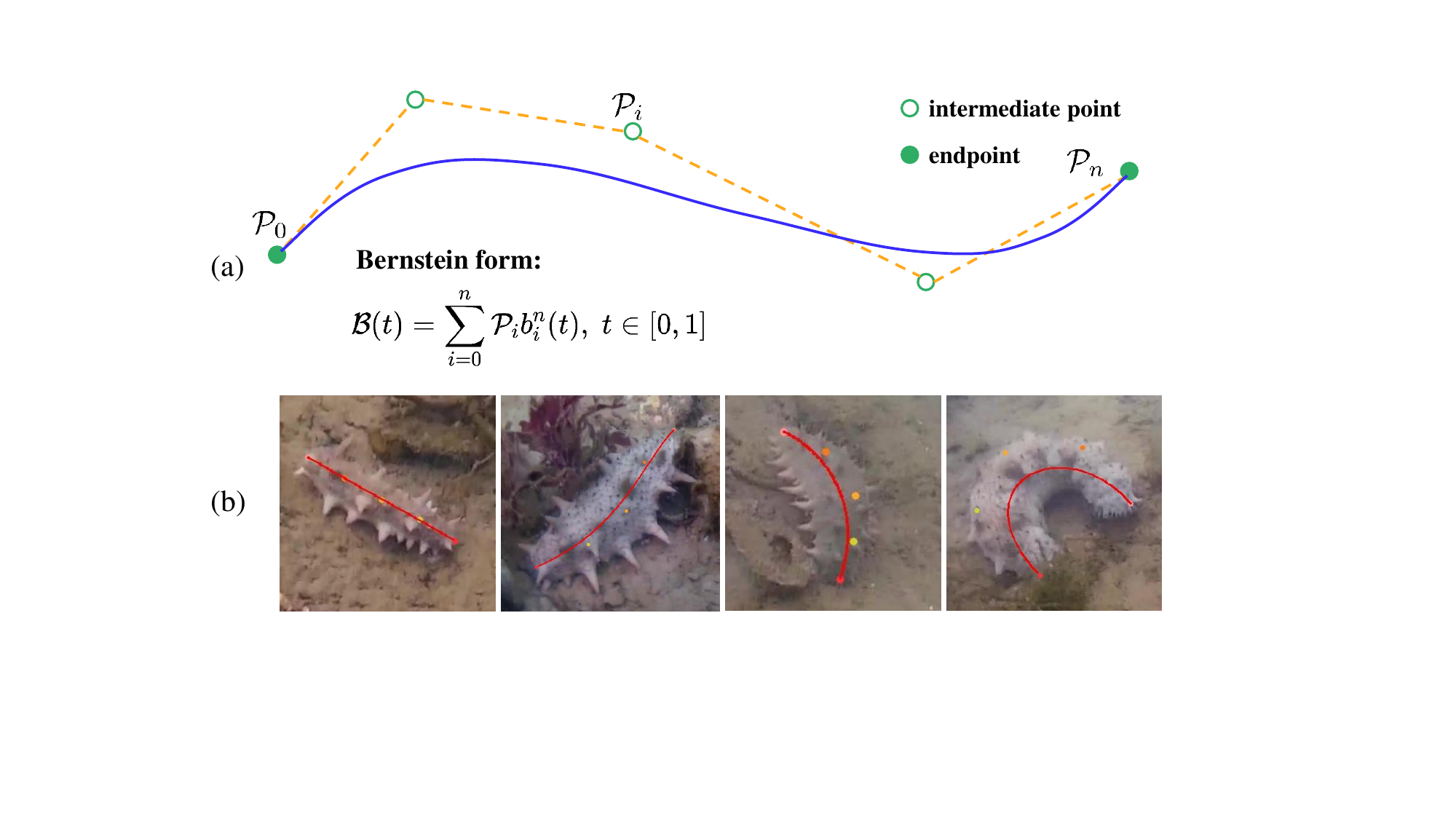}
   \caption{The B\'{e}zier curve with control points and its application on sea cucumber trunk modelling. (a) The formulation and visualization of B\'{e}zier curve with control points. The solid points are endpoints, which directly determine the length of the curve. The hollow points are intermediate points, which affect the curvature of the curve. $\mathcal{B}(t)$ is the Bernstein form of B\'{e}zier curve. $b_{i}^{n}(t)$ are Bernstein scalar polynomials of degree $n$ (see Eq.~\eqref{Bernstein_scalar}). (b) Sea cucumber trunks in various poses modelled with $4$th order B\'{e}zier curves.}
   \label{fig:Bezier}

\end{figure}

\subsection{B\'{e}zier Curve and the Sea Cucumber Trunk Modelling}\label{sec:curve modelling}
In its natural state, sea cucumber exhibits a wide range of pose variations, necessitating the utilization of a modelling technique capable of accurately capturing these diverse postures. Considering the characteristic smooth and curved nature of the sea cucumber trunks, it is natural to employ B\'{e}zier curves for modeling purposes, as they have demonstrated successful applications in lane detection \citep{Rethinking_Bezier_curve} and scene text detection tasks \citep{Liu2022ABCNetv2}.

The B\'{e}zier curve is a fundamental tool for computer graphics image modeling, and one of the most commonly used basic elements in graphic modelling. The B\'{e}zier curve fitting process based on the control points is shown in Fig. \ref{fig:Bezier}. It creates and edits smooth curve graphics by controlling the points on the curve, including intermediate points and endpoints (i.e., the start and end point). The Bernstein form of an $n$-th order B\'{e}zier curve, which is defined by $n+1$ control points, is shown in Eq. \eqref{Bezier_equation}, 

\begin{equation}
\label{Bezier_equation}
\mathcal{B}(t)=\sum_{i=0}^{n}\mathcal{P}_{i}b_{i}^{n}(t), \; t\in\left[0,1\right],
\end{equation}
where $\mathcal{P}_{i}$ is the $i$\text{-}th control point, specifically, $\mathcal{P}_{0}$ and $\mathcal{P}_{n}$ are the two endpoints which are crucial for measuring the length of the curve. $b_{i}^{n}(t)$ are the Bernstein scalar polynomials of degree $n$, which is defined as: 
\begin{equation}
\label{Bernstein_scalar}
b_{i}^{n}(t)=\binom{n}{i}t^{i}(1-t)^{n-i}, \; i=0,\ldots,n,   
\end{equation}
where $\binom{n}{i}$ is the binomial coefficient. The B\'{e}zier curves can be approximated through polynomial fitting. The higher the order determined by the number of control points, the greater the variation in curvature, rendering curves more suitable for intricate graphical modelling. Despite the potential benefits, higher-order curves do not yield significant improvements due to their increased degree of freedom, which in turn leads to instability and a heavier computational burden \citep{Rethinking_Bezier_curve}. Therefore, to achieve a better balance between speed and accuracy, we utilize the classical $4$th-order polynomial B\'{e}zier curve ($n = 4$). As demonstrated in Fig. \ref{fig:Bezier} (b), this curve is sufficiently capable of modelling sea cucumber trunks with various pose changes, especially the trunks with large bending curvatures.

\subsection{Trunk Identification Network for Sea Cucumber}
As illustrated in the examples depicted in Figures \ref{fig:strategies} and \ref{fig:Bezier}, the sea cucumber habitat often presents intricate scenes and distinguishing the sea cucumber target from its background presents a challenge due to their similar color appearance. In addition, under the perspective of unrestricted exploration by the underwater vehicle, there exists a significant disparity in scale between the visual scene captured by the camera and the sea cucumber target, rendering direct curve detection and modelling on the entire image infeasible. 

To overcome the aforementioned challenges and achieve in-situ sea cucumber trunk identification in large-scale underwater environments, we propose a novel end-to-end joint optimizing network for simultaneous sea cucumber detection and trunk curve prediction. Sea cucumber detection, as a pivotal task, serves as a means to bridge the scale disparity between the vast visual scene and the minute sea cucumber target. Beyond that, as a learning task with low annotation requirements, object detection can effectively discriminate the sea cucumber targets from the complex background and improve the accuracy of curve detection and modelling. 
\begin{figure*}[t]
\centering
	\includegraphics[width=0.98\linewidth]{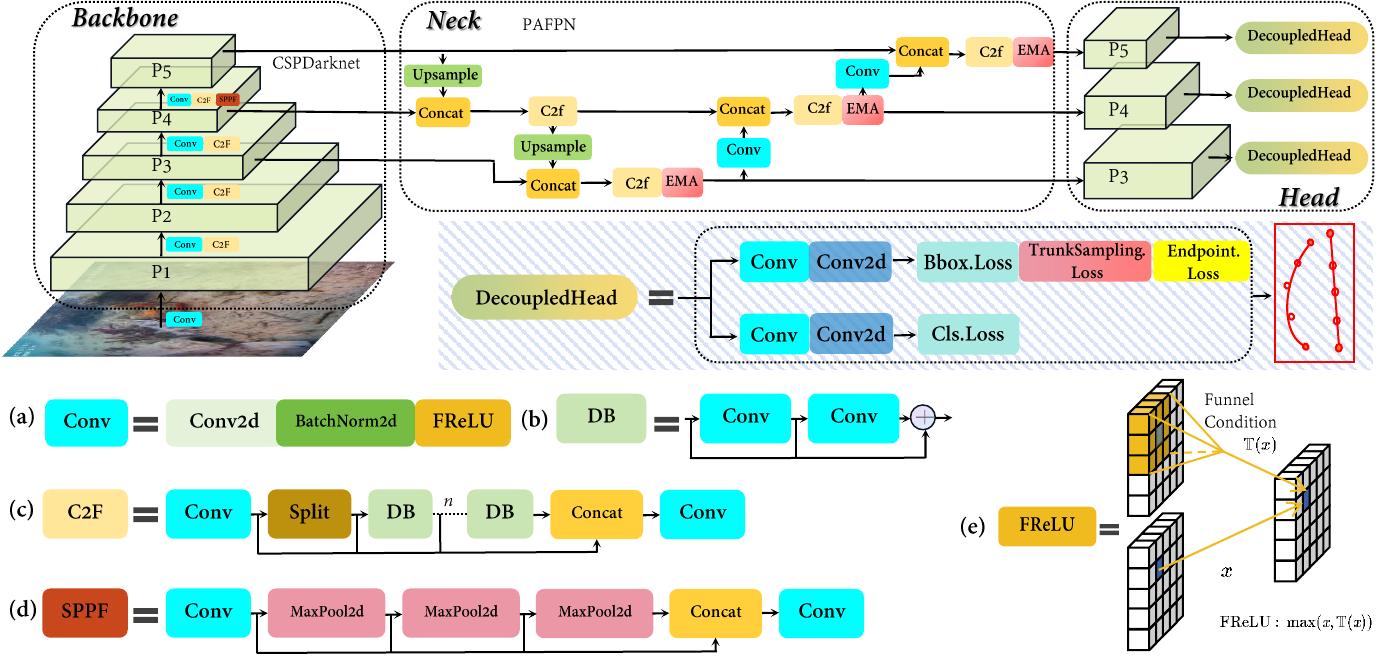}
    \caption{The proposed TISC-Net for simultaneous sea cucumber detection and trunk identification, which is an extension of YOLOv8. Feature from a CSPGarknet is strengthened by the SPPF, and then different levels of feature maps through upsampling operations are fused in the neck module. At last, the head module predicts bounding boxes and B\'{e}zier curves through a classification branch and a regression branch in one stage.}
    \label{fig:architecture}
\end{figure*}
In contrast to conventional multi-task joint learning and optimizing networks, which comprise multiple independent branches and only share weights in the backbone while disregarding task correlations, we incorporate B\'{e}zier curve modelling into the object detector for curve detection, enabling comprehensive parameter and feature sharing across multiple tasks. We adopt YOLOv8 \citep{yolov8_ultralytics}, a state-of-the-art object detection framework, as our base model and propose specific modifications tailored for trunk identification, including the seamless and lightweight integration of B\'{e}zier curve detection into the object detection framework. The architectural design of the proposed Trunk Identification Network for Sea Cucumber, based on YOLOv8\footnote{\url{https://github.com/ultralytics/ultralytics}}, referred to as TISC-Net, is depicted in Fig. \ref{fig:architecture}. From Fig. \ref{fig:architecture} (a) to (e), the detailed structure of the key composition modules, including the ConvModule (Conv), Darknet Bottleneck (DB), CSPLayer\_2Conv, Spatial Pyramid Pooling-Fast (SPPF) , and the FReLU activation, are presented.

The YOLOv8 model primarily adopts a similar backbone to that of YOLOv5, which also comprises backbone, neck, and head in its architecture. The CSPLayer has been enhanced to achieve improved contextual understanding of high-level features, with its upgraded version known as CSPLayer\_2Conv (C2f in Fig. \ref{fig:architecture}). The YOLOv8 model adopts an anchor-free strategy, which designs multi-scale decoupled heads (DecoupleHead in Fig. \ref{fig:architecture}) to enable the independent processing of classification and regression tasks, distinguishing it from previous anchor-based strategies. For sea cucumber detection, given the rich target representation in the features utilized for bounding box localization within the regression branch, a straightforward, cost-effective, and seamless approach to merging object detection and curve modelling tasks is to directly modify the composition of output from said regression branch. In our TISC-Net, the regression branch of the multiple decoupled heads are expanded to incorporate the prediction of parameters for B\'{e}zier curve modelling, thereby the composition of the new regression outputs are as follows:
\begin{itemize}
\setlength{\itemsep}{-1pt}
    \item[•] the original $\textbf{4}$ regression outputs for the bounding box of the sea cucumber target, encompassing the plane coordinates of its center, width, and height;
    \item[•] and another $\textbf{10}$ outputs for the trunk modelling with a B\'{e}zier curve, i.e. the coordinates of its $5$ control points.
\end{itemize} 
The classification branch, which is another component of the decoupled head, remains unchanged in its original form as seen in YOLOv8, with a single output indicating the presence of the sea cucumber.

The detection of the sea cucumber target itself is already a challenging problem, considering the interference from complex backgrounds and the similarity in appearance to the background. The introduction of the trunk identification task further complicates this joint learning problem. The detection and modelling of curves for trunk identification impose higher demands on the feature representation of trunk spatial context and cross-spatial learning capability. Based on this analysis, we employ advanced feature activation technology and attention mechanism to further enhance the backbone and neck structure in the aforementioned YOLOv8 model, rendering it more tailored for our specific target task. More details will be given in the following Section \ref{sec:curvefeat}.

\subsection{Improved Curve Feature Perception and Learning}\label{sec:curvefeat}
Given the resemblance between sea cucumber target and background, local context and cross-spatial features play crucial roles in trunk identification and modelling tasks compared to other general target perception tasks. The lack of robustness in the feature representation for capturing spatial context awareness often leads to a phenomenon where control points are erroneously positioned in the background region, thereby causing failure in skeleton recognition.

\subsubsection{Funnel Activation for Curve Contextual Perception}\label{sec:FReLU}
To enhance the local context representation ability for describing sea cucumber target features, we substitute the SiLU activation in the YOLO framework with the funnel activation (FReLU) \citep{FReLUl}, which exhibits superior spatial context perception capabilities. Ma et al. contend that the limited responsiveness of the activation function hinders further advancements in visual perception. Therefore, unlike traditional activation functions that depend solely on the pixel value, FReLU expands ReLU and PReLU to 2D activation by adding a visual funnel condition $\mathbb{T}(\cdot)$ defined on the local context, as illustrated in Fig. \ref{fig:architecture} (e). The funnel-shaped condition enables the network to create spatial conditions through the nonlinear activation of each pixel. The integration of context-aware activation in our sea cucumber trunk identification task enhances the sensitivity of TISC-Net for challenging targets, enabling it to achieve robust and comprehensive spatial features while only introducing a marginal increase in parameters.

\subsubsection{Curve Learning with Cross-Spatial Dependency}\label{sec:EMA}
To further enhance cross-channel interaction and establish cross-spatial dependency for the features utilized in simultaneous object detection and curve prediction, we integrate the efficient multi-scale attention (EMA) \citep{EMA_attention} into the Neck component of the TISC-Net. The EMA model aims to acquire effective channel descriptions without reducing the dimensionality of channels, resulting in a superior pixel-level attention mechanism for high-level feature maps.
The EMA first groups the input feature map into sub-features based on channel dimensions in order to ensure well-distributed semantics within each group. Then, three parallel routes, i.e., two $1\times 1$ convolution-based branch and one $3\times 3$ convolution-based branch, are designed to extract attention weight descriptors. Consequently, by employing 1D global average-pooling for encoding along the horizontal dimension, EMA effectively captures long-range dependencies in the horizontal direction while preserving precise positional information in the vertical direction.

\subsection{Loss Function for Multiple Task Joint Learning}
Our TISC-Net is a multi-task joint learning framework designed for object detection and point prediction-based trunk identification, with its training heavily reliant on an effective loss function. While the object detection task already has a well-defined and proven effective loss function, the true challenge in designing a loss function for TISC-Net lies in accurately quantifying the deviation between the ground truth curve and predicted one. 

The B\'{e}zier curve modelling, as depicted in Fig. \ref{errordeviation} (a), exhibits both robustness and fragility in our trunk identification and length measurement task. The robustness of the B\'{e}zier curve is demonstrated by its ability to withstand minor deviations at intermediate points without compromising overall integrity; however, even a slight deviation at the endpoint can have fatal consequences for our length measurement task (Fig. \ref{errordeviation} (b)). This finding suggests that attention should be paid not only to the degree of fit between the predicted trunk and ground truth, but also to the deviation of the predicted endpoints. Therefore, we introduce a combined loss for trunk identification, which includes a sampling loss term and an endpoint loss term.

\subsubsection{Trunk Sampling Loss}
\begin{figure}[t]
\centering
\includegraphics[width=1\linewidth]{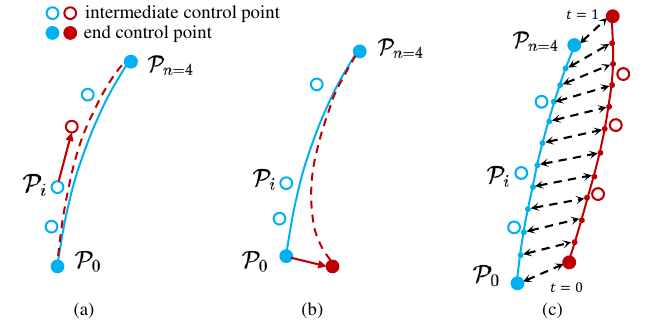}
\caption{The trunk sampling loss. (a) The overall integrity of a B\'{e}zier curve is rarely affected by slight deviations at the intermediate points. (b) The location of the entire B\'{e}zier curve is significantly compromised by even a slight deviation in its endpoints (end control points). (c) The trunk sampling loss for measuring the curve deviation.}
\label{errordeviation}
\vspace{-1em}
\end{figure}

For a curve modeling task, the curve location loss aims to precisely quantify the discrepancy between the ground truth and predicted curves. Inspired by the works of \citep{Rethinking_Bezier_curve} and \citep{Liu2022ABCNetv2}, it is evident that the dissimilarity between two curves cannot be adequately captured solely by employing the $L1$ loss on their control points. Therefore, we adopt a more appropriate curve loss as proposed in \citep{Rethinking_Bezier_curve} (Fig.~\ref{errordeviation} (c)), namely sampling loss, which samples curves at uniformly spaced points $t\in \left [ 0,1 \right ] $ and measures the distance between corresponding sampling points from two curves. The $t$ values can be either an ordered increasing series or further transformed by re-parameterizing the function $f\left ( t \right )$ to enhance the sampling strategies. Specifically, the trunk sampling loss $\mathcal {L}_{tsl}$ is depicted in Eq. (\ref{eq:TSL_equation}):
\begin{equation}
\label{eq:TSL_equation}
\mathcal {L}_{tsl} =\frac{1}{n} \sum_{t\in T}\left \| \mathcal{B} \left ( f\left ( t \right )  \right ) -  \hat{\mathcal{B}}  \left ( f\left ( t \right )  \right ) \right \|_{1},  
\end{equation}
where $\mathcal{B}\left ( t \right ) $ is the ground truth and $\hat{\mathcal{B}}\left ( t \right ) $ is the predicted curve, 
the variable $n$ denotes the number of samples in the sample set $T$ and has been empirically set to 50. We adopt an equally spaced sampling strategy, commonly employed in other curve-based tasks~\citep{tabelini2021polylanenet, Rethinking_Bezier_curve}, where $f(t)=t$. The adoption of this sampling loss equally emphasizes the local curve throughout the entire trunk, thereby facilitating model convergence and enhancing its generalization capabilities.

After specifying the sampling strategy as $f(t)=t$, it is necessary to generate the ground truth of the B\'{e}zier curve. This process can be divided into two distinct steps. Firstly, we follow \citep{Rethinking_Bezier_curve} and ~\citep{Liu2022ABCNetv2} to determine the specific B\'{e}zier curve function (Eq.~\eqref{Bezier_equation}) by fitting the annotation points $\left \{ \left ( p_{x_{i}}, p_{y_{i}} \right )  \right \} _{i=0}^{m} $ using standard least squares, where $\left ( p_{x_{i}}, p_{y_{i}} \right )$ denotes the coordinates of the $i$\text{-}th point. Then, in conjunction with the sampling strategy and the number of samples, the control points $\left \{ \mathcal{P} _{i} \left ( x_{i}, y_{i} \right )  \right \} _{i=0}^{n} $ can be easily calculated. For clarity, the matrix expression that generates the ground truth of the B\'{e}zier curve-based trunk is given by Eq. (\ref{eq:Curve_Ground_Truth_Generation}):
\begin{equation}
\label{eq:Curve_Ground_Truth_Generation}
\begin{bmatrix}\mathcal{P}_{0} 
 \\\mathcal{P}_{1} 
 \\\vdots 
 \\\mathcal{P}_{n} 
\end{bmatrix}=
\begin{bmatrix}
  b_{0}^{m}(t_{0}) & \cdots  &  b_{m}^{m}(t_{0})\\
  b_{0}^{m}(t_{1}) & \cdots  &  b_{m}^{m}(t_{1}) \\
 \vdots& \ddots   &  \vdots \\
  b_{0}^{m}(t_{n}) & \cdots  &  b_{m}^{m}(t_{n})
\end{bmatrix}
\begin{bmatrix}
  p_{x_{0}} & p_{y_{0}}   \\
  p_{x_{1}} & p_{y_{1}}  \\
 \vdots & \vdots \\
  p_{x_{m}} & p_{y_{m}}  
\end{bmatrix},
\end{equation}
where $\left \{ t_{i}  \right \} _{i=0}^{n}$ is equidistant sequence from 0 to 1. We follow \citep{Rethinking_Bezier_curve} without limiting the number of original annotations, resulting in increased degrees of freedom and expandability.
\subsubsection{Trunk Endpoint Loss} \label{sec:endpoint_loss}
The importance of endpoints for Bézier curve-based trunking modeling and length measurement is demonstrated in Fig.~\ref{errordeviation} (b). An endpoint loss is therefore introduced to impose a stricter penalty on deviations between the true endpoints and their predicted ones. The general loss functions for key points regression are $L1$, $L2$ or $smooth L1$. However, it is found that these loss terms are not sensitive to small errors. To address this issue, the wing loss ~\mbox{\citep{feng2018wingloss}} is introduced to impose an additional robust and sensitive constraint on the deviation of endpoints, i.e.,
\par
\begin{equation}
\label{wingloss_equation}
wing \left ( x \right ) =\left\{\begin{matrix}w\cdot ln \left ( 1+  \left | x \right |/\epsilon  \right ),\enspace\mathrm{if}\enspace \left | x \right |< w   \\\left | x \right | - C,\qquad  \qquad \mathrm{otherwise}\end{matrix}\right. ,
\end{equation}
where the non-negative parameter $w$ determines the range of the non-linear component, which is bounded within $\left ( -w,w \right ) $, the parameter $\epsilon$ controls the curvature of the non-linear region. Additionally, a constant $C=w-wln\left ( 1+w/\epsilon \right ) $ smoothly connects the piecewise-defined linear and non-linear parts. We adhere to the default parameters as reported in \citep{feng2018wingloss}. Then, the trunk endpoint loss is defined as:
\begin{equation}
\label{endpointloss_equation}
\mathcal {L}_{epl} =\sum_{i\in \left \{ 0, n\right \}} wing\left (\mathcal{D} \left(\mathcal{P}_{i}, \mathcal{P}'_{i}  \right ) \right ) ,
\end{equation}
where $\mathcal{P}_{i}$ and $\mathcal{P}'_{i}$, $i\in \left \{ 0, n\right \} $ represent the predicted and ground truth endpoints, $\mathcal{D} \left(\mathcal{P}_{i}, \mathcal{P}'_{i}  \right )$ calculates their distance.  

\subsubsection{Overall Loss Function}
The above mentioned loss terms for Bézier curve-based trunk identification, along with the general object detection loss $\mathcal {L}_{det}$ from the YOLOv8 detector, contribute to the overall multi-task learning loss for our TISC-Net as follows:
\begin{equation}
\label{overall_loss}
\mathcal {L}=\lambda _{1}\cdot\mathcal {L}_{det}+\lambda _{2}\cdot  \mathcal {L}_{tsl}+ \lambda _{3} \cdot \mathcal {L}_{epl} .
\end{equation}
where $\lambda _{i}, i\in \left \{ 1, 2, 3 \right \}$ are weight factors for different loss terms. The values of $\lambda _{1}$, $\lambda _{2}$, and $\lambda _{3}$ were empirically set to 1, 1, and 0.1 in our experiments.


\subsection{Length Measurement via 3D Curve Integration}
The proposed TISC-Net enables the convenient acquisition of sea cucumber trunks, which can then be combined with the 3D information provided by the binocular camera to accurately measure the length of the sea cucumber trunk using a curve integration method defined in the 3D underwater environment. The length measurement via 3D curve integration can be expressed by the parametric equation, as illustrated in Eq. \eqref{integration_equation}.
\begin{equation}
\label{integration_equation}
Length = \int _{\mathbb{L} } \sqrt{(x_{(t)}')^{2}    + (y_{(t)}')^{2} + (z_{(t)}')^{2}} \mathrm{d}t,
\end{equation}
where $t$ represents the local position where the curve $\mathbb{L}$ passes, while $x, y, z$ denote the 3D coordinates in the Cartesian coordinate system. In this problem, the discrete form of Eq. \eqref{integration_equation} can be written as
\begin{equation}
\label{discrete}
Length =\sum_{m=0}^{M} \sqrt{ \sum_{i\in\{x,y,z\}}^{}\left (P_{m+1,i}- P_{m,i}\right )^2},
\end{equation}
where $m$ is the index of the $M+1$ 3D points that the curve $\mathbb{L}$ passes through in the point cloud.

In our sea cucumber harvesting robot, a ZED series binocular camera developed by Stereolabs was utilized to acquire pixel-level 3D information, leveraging its integrated stereo vision technology-based scene depth recovery algorithm. The ZED camera, unlike depth cameras based on Time of Flight (TOF) and structured light, does not rely on projecting a light source onto the external active light. Instead, it solely relies on the polar constraints of the two captured images to generate a depth map and its corresponding 3D point cloud. The main parameters of the adopted ZED camera are available on the StereoLabs homepage\footnote{\url{https://www.stereolabs.com/products/zed-2}}. The ZED camera underwent waterproofing and recalibration procedures before being installed on our harvesting robot.

\section{Experiments}

\subsection{Benchmarks}
To assess the proposed method, we conducted experiments on two benchmark datasets: one being the newly established Sea Cucumber In-situ Trunk Identification (\textbf{SC-ISTI}) dataset, and the other being an extension of a dataset originally designed for underwater object detection, known as Sea Cucumber from Detecting Underwater Objects (\textbf{SC-DUO}). We provide weak annotation for the trunk identification task. The additional statistics of these two benchmark datasets are presented in Table~\ref{tab:Dataset}, while more detailed information and labelling strategy can be found in the subsequent subsections.

\begin{figure*}[t!]
\centering
	\includegraphics[width=0.98\linewidth]{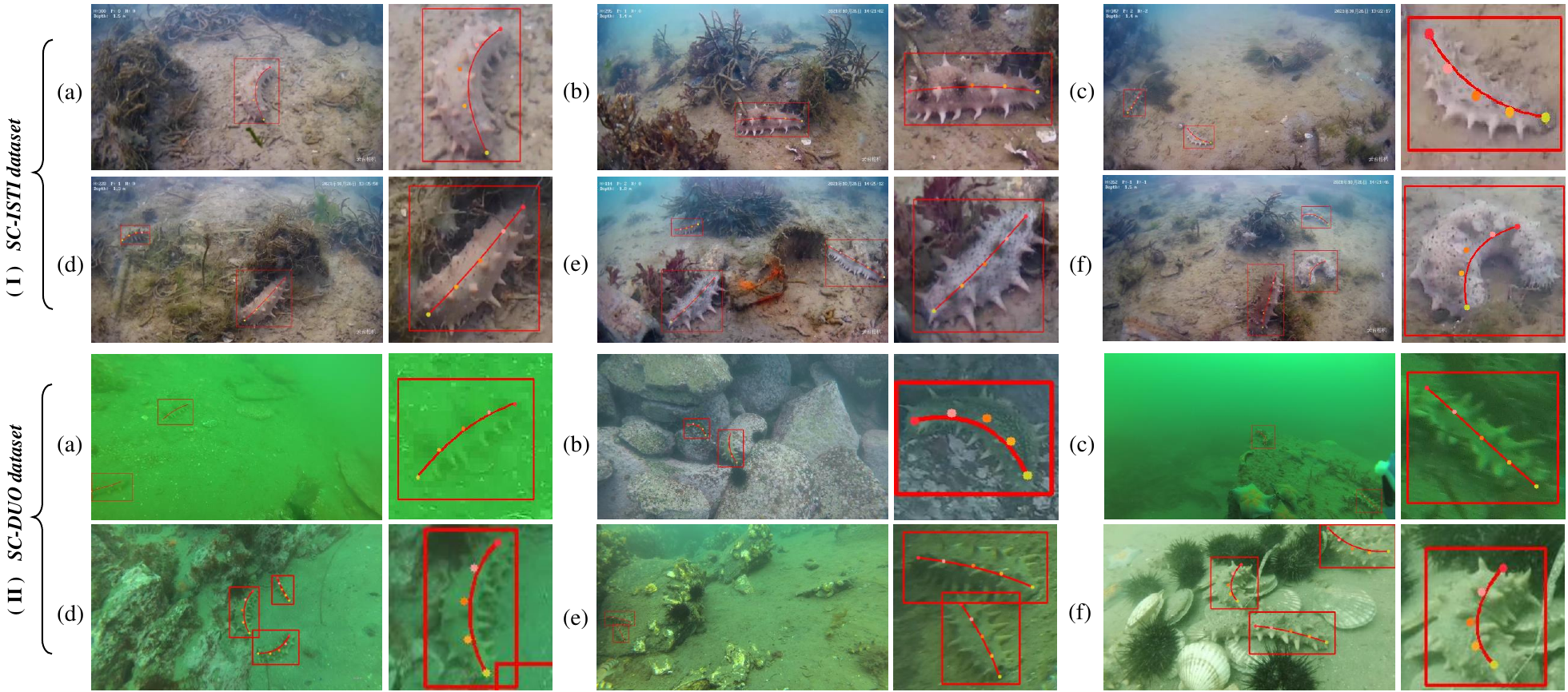}
    \caption{Examples of trunk identification with the proposed TISC-Net. Outcomes of TISC-Net-m are presented. Each image is followed by the enlarged slice of the target, with B\'{e}zier curve control points marked by solid circles. The images have been slightly resized for alignment.}
    \label{Fig:datasets_example}

\end{figure*}

\begin{table}[t]
\caption{Details of the SC-ISTI and SC-DUO datasets. $N_{img}$ represents the number of images, while $N_{sc}$ denotes the quantity of sea cucumbers. \label{tab:Dataset}} 
\fontsize{9pt}{12pt}\selectfont
\centering
\begin{tabular}{ C{0.15\linewidth} | C{0.08\linewidth} C{0.08\linewidth} | C{0.08\linewidth} C{0.08\linewidth}| C{0.18\linewidth} }
\Xhline{1pt}
\multirow{2}{*}{\textbf{Dataset}} & \multicolumn{2}{c|}{\textbf{Train/Validation}} & \multicolumn{2}{c|}{\textbf{Test}} & \multirow{2}{*}{\textbf{Resolution}} \\ \cline{2-5}
 & $N_{img}$ & $N_{sc}$  & $N_{img}$ & $N_{sc}$ &  \\ 
 \hline
SC-ISTI & 370 & 526  & 92 & 134 & 1920$\times$1080 \\
SC-DUO  & 879 & 1216 & 144 & 207 & 720$\times$405 \\
\Xhline{1pt}
\end{tabular}
\vspace{-1em}
\end{table}

\subsubsection{SC-ISTI Dataset}
The SC-ISTI dataset comprises $462$ RGB images captured by an underwater robot in real habitat of sea cucumbers, with a resolution of $1920\times 1080$. These images were acquired in October 2021 at the marine ranch situated in Weihai, Shandong Province, China. Among these $462$ images, there are a total of $670$ sea cucumbers present, which have been randomly divided into training, validation and testing sets. The training/validation set totally consists of $370$ images containing $526$ sea cucumbers, while the testing set comprises $92$ images with a count of $134$ sea cucumbers. Several illustrative examples from this dataset are depicted in Fig.~\ref{Fig:datasets_example}. The intricate benthic environment poses significant disruption and presents great challenges for vision-based tasks.

\subsubsection{SC-DUO Dataset}
The DUO dataset \citep{DUOliu2021dataset} is a comprehensive benchmark for underwater object detection, comprising $7,782$ carefully curated and re-annotated images with a resolution of $720\times 405$. It encompasses four marine creature categories, namely holothurian (sea cucumber), echinus, scallop, and starfish. The SC-DUO dataset is created by selecting images containing sea cucumbers and adding corresponding trunk labels, making it suitable for the trunk identification task. Specifically, the SC-DUO dataset consists of $1,023$ images containing $1,856$ sea cucumbers. To facilitate training and evaluation processes, we further partitioned this dataset into training/validation and testing sets. The training/validation set comprises $879$ images with $1,216$ sea cucumbers while the testing set contains $144$ images with $207$ sea cucumbers. Notably, compared to the SC-ISTI Dataset, the SC-DUO dataset offers more diverse underwater scenes captured under challenging imaging conditions (refer to Fig.~\ref{Fig:datasets_example}).

\subsubsection{Weak Annotation for Trunk Identification}
Taking advantage of the superior curve modelling capability of B\'{e}zier curves, we can avoid the need for pixel-annotated masks to label sea cucumber trunks and instead rely on a few key points along the curves. This weak annotation method is cost-effective yet fully descriptive of the curves, requiring only the sea cucumber bounding box and trunk key points. It should be noted that the number of key points can be adjusted as needed; however, given the diversity of trunks, it is a reasonable choice to use five key points as discussed in Section \ref{sec:curve modelling}.

\subsection{Experiment Settings and Implementation Details}

The proposed TISC-Net was implemented on the PyTorch platform, encompassing models of various scales: an extra large-size model (xP6), large-size models (x), medium-size models (l, m), and small-size models (n, s). The postfix ``6" in the model name indicates the inclusion of the P6 output block in the HeadModule. These models all utilize YOLOv8 CSPDarknet as their backbone, with variations in depth multiples (D), width multiples (W), and maximum channels (C). Beyond that, the Parameters, FLOPS, and FPS for each of these models are also listed in Table~\ref{tab:Parameters}.

We conduct training for $250$ epochs on both SC-ISTI and SC-DUO datasets, using a batch size of $16$. For optimization, we employ the Adam optimizer with a learning rate of $1e-3$ and weight decay of $1e-4$. Additionally, we adopt the Cosine Annealing learning rate schedule as described in~\citep{tabelini2021keep}. All hyper-parameters remain consistent across different datasets. Our data augmentation techniques include random cropping, random horizontal flips, and Mosaic. The experiments are conducted on a consistent hardware setup comprising an NVIDIA RTX 3090 GPU, a $2.3$GHz Intel Xeon processor, $128$GB RAM, and the Ubuntu 18.04 operating system platform.

\begin{table*}[]
 \caption{Details of the TISC-Net models, where (D, W, C) are the depth multiples, width multiples and max channels of the CSPDarknet. The number of parameters, FLOPS and FPS of all TISC-Net models and the compared methods are also reported. \label{tab:Parameters}} 
 \fontsize{9pt}{10pt}\selectfont
 \renewcommand{\arraystretch}{1.3}  
 \setlength\tabcolsep{3pt}
 \centering
 \scalebox{1}{
   \begin{tabular}{ C{0.15\linewidth} | C{0.15\linewidth} | C{0.15\linewidth} | C{0.1\linewidth} | C{0.1\linewidth} | C{0.08\linewidth} | C{0.15\linewidth} }
 \Xhline{1pt}
 \textbf{Model} & \textbf{Backbone} & \textbf{(D, W, C)} & \textbf{Params (M)} & \textbf{FLOPs (B)} & \textbf{FPS} & \textbf{Reference} \\ 
 \hline
 TISCNet-n & CSPDarknet & (0.33, 0.25, 1024) & 3.1 & 9.1 & 110 & \textbf{-}  \\
 TISCNet-s & CSPDarknet & (0.33, 0.50, 1024) & 11.4 & 30.3 & 83 & \textbf{-} \\
 TISCNet-m & CSPDarknet & (0.67, 0.75, 768)  & 26.4 & 81.5 & 66 & \textbf{-} \\
 TISCNet-l & CSPDarknet & (1.00, 1.00, 512)  & 44.5 & 169.1 & 50 & \textbf{-}\\
 TISCNet-x & CSPDarknet & (1.00, 1.25, 512)  & 69.5 & 263.7 & 43 & \textbf{-} \\
 TISCNet-x-P6 & CSPDarknet & (1.00, 1.25, 512)  & 99.1 & 1066.8 & 38 & \textbf{-}  \\ 
 \hline
 \begin{tabular}[c]{@{}c@{}}FasterRCNN\\ +RTMpose\end{tabular} & \begin{tabular}[c]{@{}c@{}}ResNet50\\ CSPNeXt\end{tabular} & -                & \begin{tabular}[c]{@{}c@{}}41.3\\ 27.6\end{tabular}           & \begin{tabular}[c]{@{}c@{}}213.0\\ 4.1\end{tabular}          & 11 & \begin{tabular}[c]{@{}c@{}}\citep{ren2015faster}\\ \citep{jiang2023rtmpose}\end{tabular}           \\ 
 \hline
 \begin{tabular}[c]{@{}c@{}}RTMdet\\ +RTMpose\end{tabular}     & \begin{tabular}[c]{@{}c@{}}CSPNeXt\\ CSPNeXt\end{tabular}  & -                & \begin{tabular}[c]{@{}c@{}}52.3\\ 27.6\end{tabular}           & \begin{tabular}[c]{@{}c@{}}79.9\\ 4.1\end{tabular}           & 13 & \begin{tabular}[c]{@{}c@{}}\citep{lyu2022rtmdet}\\ \citep{jiang2023rtmpose}\end{tabular}           \\ 
 \Xhline{1pt}
 \end{tabular}
 }

 \end{table*}
\subsection{Comparison}
\subsubsection{Compared Methods}
Considering that curve detection under large-scale wild scenarios itself is a novel problem, there lacks well-established frameworks for comparative analysis. Therefore, in addition to employing TISC-Net with different scales as the comparison models, we further introduce two integrated two-stage top-down frameworks, i.e., \textbf{Faster-RCNN} \citep{ren2015faster} + \textbf{RTMpose} \citep{jiang2023rtmpose}, and \textbf{RTMdet} \citep{lyu2022rtmdet} + \textbf{RTMpose} \citep{jiang2023rtmpose}.

The Faster-RCNN and RTMdet models are two representative deep object detectors, in addition to the YOLO series. RTMpose is an advanced pose estimation framework that incorporates a key-point localization module and treats it as a classification task. These two integrated two-stage top-down frameworks initially generate bounding boxes, followed by uniform scaling of the sea cucumbers for subsequent key-point (control point) detection of the Bézier curve. For clarity, the details of these two compared methods are also listed in Table \ref{tab:Parameters}. These methods are implemented on the OpenMMLab platform~\citep{sengupta2020mm}. The two compared models were trained on both the SC-ISTI dataset and the SC-DUO dataset. All of these methods, including our proposed models, utilized identical training/validation and test sets.

\subsubsection{Evaluation Metrics}
The YOLO officially adopted metrics, such as mAP50 and mAP50-95, can be utilized for the quantitative evaluation of sea cucumber object detection. Considering that the sea cucumber trunks are annotated by points on the B\'{e}zier curve, the correspondence between these points can be employed to represent the alignment between the predicted and ground truth curves. By sampling 50 equidistant points on-curve, we assess the point correspondences using official COCO Metrics to evaluate the curve alignments, including mAP and Percentage of Correct Keypoints (PCK). More details about the COCO Metrics are available on its project home page\footnote{\url{https://cocodataset.org/\#keypoints-eval}}.

\subsubsection{Overall Performance Comparison and Discussion}
\begin{figure}[t]
\centering
	\includegraphics[width=1\linewidth]{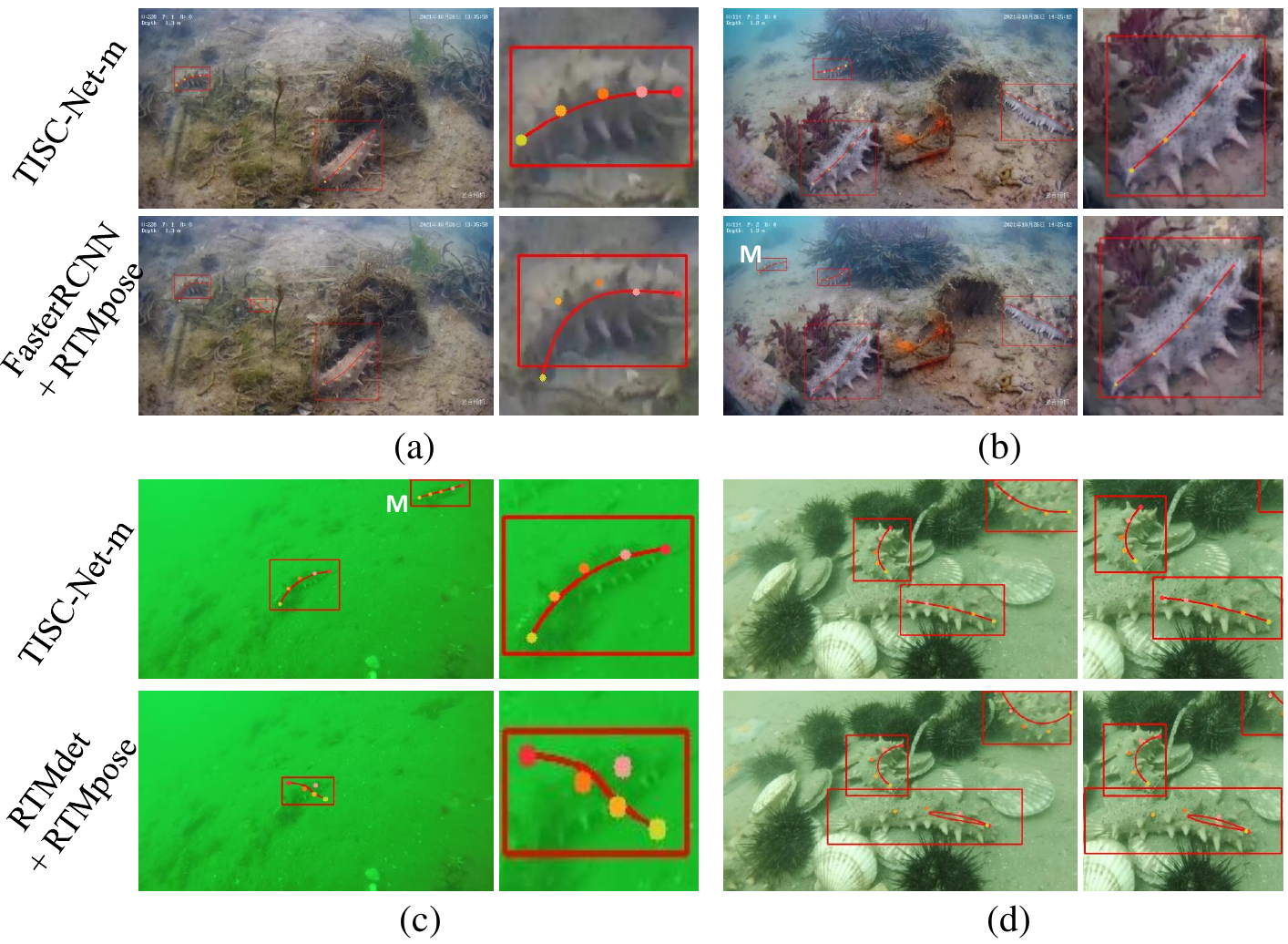}
    \caption{Visual comparison of TISC-Net-xP6 and the compared two-Stage methods on the test sets of SC-ISTI and SC-DUO. The two-stage(\textcolor{red}{\textbf{1}}) method is the FasterRCNN\citep{ren2015faster}+RTMpose\citep{jiang2023rtmpose}, and the two-stage(\textcolor{red}{\textbf{2}}) method is the RTMdet\citep{lyu2022rtmdet}+RTMpose\citep{jiang2023rtmpose}. The white ``M'' character indicates that a target is missing for the other method. All other markup information is consistent with Fig.~\ref{Fig:datasets_example}.}
    \label{Fig:comparison_of_methods}
\vspace{-1em}
\end{figure}

The quantitative comparisons on the test set of SC-ISTI and SC-DUO datasets are reported in Table \ref{tab:comparison}. In addition to evaluating the performance of trunk identification, the performance of sea cucumber detection is also assessed. Fig. \ref{Fig:datasets_example} presents examples of trunk identification with the proposed TISC-Net, where both the detected bounding boxes and the trunk curves are plotted. 
 The visual examples of the comparison between the proposed TISC-Net and the combined two-stage methods on the two test datasets is illustrated in Fig. \ref{Fig:comparison_of_methods}. 
 
\begin{table*}[t]
\caption{Results on the $\mathit{test}$ set of SC-ISTI and SC-DUO. The two-stage methods reproduced results in the OpenMMLab platform, best performance from three random runs. The top three scores are in \textcolor{red}{red}, \textcolor{green}{green}, \textcolor{blue}{blue}. (Best viewed in color)\label{tab:comparison}}  
\fontsize{11.5pt}{12pt}\selectfont
\setlength\tabcolsep{2pt}
\renewcommand{\arraystretch}{1.5}  
\centering
\resizebox{1.0\linewidth}{!}{

\begin{tabular}{cccccccccccccl}
\hline
                                   & \multicolumn{6}{c}{\textbf{SC-ISTI}}                                                                                                                        &  & \multicolumn{6}{c}{\textbf{SC-DUO}}                                                                                                                         \\ \cline{2-7} \cline{9-14} 
                                   & \multicolumn{2}{c}{\textbf{Object Detection}}               &  & \multicolumn{3}{c}{\textbf{Trunk Identification}}                                                  &  & \multicolumn{2}{c}{\textbf{Object Detection}}               &  & \multicolumn{3}{c}{\textbf{Trunk Identification}}                                                  \\
\multirow{-3}{*}{\textbf{Methods}} &\scalebox{0.9}{\textbf{mAP50}$\uparrow$ }               & \scalebox{0.9}{\textbf{mAP50-95}$\uparrow$}            &  & \scalebox{0.9}{\textbf{mAP50}$\uparrow$ }              & \scalebox{0.9}{\textbf{mAP50-95}$\uparrow$}             & \scalebox{0.9}{\textbf{PCK}$\uparrow$ }                  &  & \scalebox{0.9}{\textbf{mAP50}$\uparrow$ }              & \scalebox{0.9}{\textbf{mAP50-95}$\uparrow$}             &  & \scalebox{0.9}{\textbf{mAP50}$\uparrow$ }               & \scalebox{0.9}{\textbf{mAP50-95}$\uparrow$}             & \scalebox{0.9}{\textbf{PCK}$\uparrow$ }                \\ \cline{1-3} \cline{5-7} \cline{9-10} \cline{12-14} 
TISC-Net-n                          & 0.920                        & 0.593                        &  & 0.858                        & 0.770                        & 0.853                        &  & 0.601                        & 0.255                        &  & 0.583                        & 0.479                        & 0.607                        \\
TISC-Net-s                          & 0.943                        & {\color[HTML]{3531FF} 0.652} &  & 0.884                        & 0.826                        & 0.866                        &  & 0.711                        & {\color[HTML]{3531FF} 0.307} &  & {\color[HTML]{3531FF} 0.703} & 0.595                        & 0.673                        \\
TISC-Net-m                          & 0.929                        & 0.591                        &  & {\color[HTML]{333333} 0.890} & 0.828                        & {\color[HTML]{3531FF} 0.920} &  & 0.695                        & 0.285                        &  & 0.679                        & 0.578                        & 0.713                        \\
TISC-Net-l                          & 0.938                        & 0.610                        &  & 0.894                        & {\color[HTML]{333333} 0.831} & 0.905                        &  & 0.655                        & 0.270                        &  & {\color[HTML]{333333} 0.652} & {\color[HTML]{3531FF} 0.562} & 0.723                        \\
TISC-Net-x                          & {\color[HTML]{3531FF} 0.956} & 0.629                        &  & {\color[HTML]{FE0000} 0.923} & {\color[HTML]{34FF34} 0.859} & 0.919                        &  & 0.682                        & 0.302                        &  & {\color[HTML]{333333} 0.692} & {\color[HTML]{34FF34} 0.611} & {\color[HTML]{34FF34} 0.745} \\
TISC-Net-xP6                        & {\color[HTML]{34FF34} 0.961} & {\color[HTML]{34FF34} 0.653} &  & {\color[HTML]{34FF34} 0.914} & {\color[HTML]{FE0000} 0.871} & {\color[HTML]{FE0000} 0.941} &  & {\color[HTML]{34FF34} 0.772} & {\color[HTML]{34FF34} 0.335} &  & {\color[HTML]{34FF34} 0.738} & {\color[HTML]{FE0000} 0.658} & {\color[HTML]{FE0000} 0.776} \\ \hline
\scalebox{0.9}{Faster-RCNN+RTMpose}                 & 0.925                        & 0.613                        &  & 0.887                        & 0.812                        & 0.913                        &  & {\color[HTML]{3531FF} 0.744} & 0.306                        &  & 0.668                        & 0.479                        & 0.691                        \\
\scalebox{0.9}{RTMdet+RTMpose}                     & {\color[HTML]{FE0000} 0.967} & {\color[HTML]{FE0000} 0.677} &  & {\color[HTML]{3531FF} 0.901} & {\color[HTML]{3531FF} 0.849} & {\color[HTML]{34FF34} 0.921} &  & {\color[HTML]{FE0000} 0.843} & {\color[HTML]{FE0000} 0.495} &  & {\color[HTML]{FE0000} 0.747} & 0.398                        & {\color[HTML]{3531FF} 0.736} \\ \hline
\end{tabular}
}

\end{table*}

We find that our two-in-one framework (i.e., object detection and trunk identification) performs well on both the SC-ISTI dataset with a complex background and the SC-DUO dataset with severe visual degradation. The good performance is reflected in the accurate matching between the detection results and the predicted trunk curve, precise prediction of the endpoints, and proper fitting of the sea cucumber trunks under natural bending postures. Table \ref{tab:comparison} demonstrates that the combined two-stage solution, RTMdet + RTMpose, outperforms in sea cucumber detection task, while our proposed TISC-Net excels in trunk identification task. In our one-stage framework, the accuracy of trunk curve detection is directly affected by target detection and its post-processing (e.g., non-maximum suppression). Therefore, TISC-Net needs to balance between these two tasks to improve curve detection while ensuring accurate target detection. However, such trade-offs are not considered in the two-stage method, resulting in higher object detection accuracy compared to our approach. The lack of task joint optimization mechanism in the two-stage method also contributes to the potential mismatch between the results of the two tasks, as exemplified by the failure instance of RTMdet + RTMpose depicted in Fig. \ref{Fig:comparison_of_methods} (c). In contrast, benefiting from the joint optimization framework, improved perception of curve features and the new curve objective function with trunk endpoint loss, TISC-Net achieves better accuracy in identifying trunks. The results presented in Table \ref{tab:comparison} demonstrate that TISC-Net-xP6 exhibits significant advantages in terms of mAP50-95 and PCK, indicating its superior accuracy in curve prediction.

Moreover, combining the quantitative results in Table \ref{tab:comparison} and the FPS reported in Table \ref{tab:Parameters}, we can notice that our one-stage end-to-end framework achieves competitive prediction performance with a threefold advantage in operational efficiency compared to the best two-stage non-end-to-end framework. Obviously, in underwater application scenarios with high real-time requirements, such as ROV-based resource survey and harvesting, the advantages of the proposed method are more apparent. Of course, we also notice some cases of identification failure in extreme situations, such as the results shown in Fig. \ref{Fig:datasets_example} (f) for both SC-ISTI and SC-DUO. In these cases, our method does not accurately predict the endpoint and trunk curves when the sea cucumber exhibited significant curvature bending in response to stress. More discussion about this issue can be found in Section \ref{sec:discussion}.

\subsection{Ablation Study}
To validate the effectiveness of the key improvements and designs in our TISC-Net, we conducted a series of ablation studies involving the following ablated models:
\begin{itemize}
    \item \textbf{-w/o-FReLU:} replacing the FReLU activation with the SiLU activation.
    \item \textbf{-w/o-EMA:} without the efficient multi-scale attention module.
    \item \textbf{-w/o-Endpointloss:} using the trunk sampling loss and removing trunk endpoint loss.
\end{itemize}

\begin{figure*}[t]
\centering
	\includegraphics[width=0.98\linewidth]{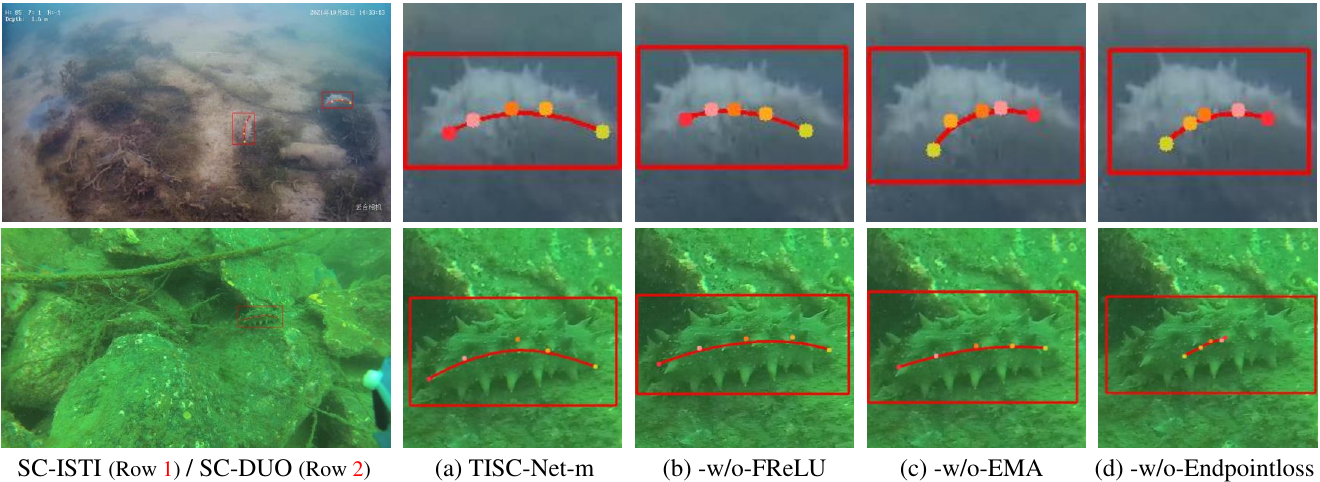}
    \caption{Exemplary results of the ablation study conducted on key modules to enhance curve feature perception and learning, encompassing FReLu activation, EMA module, as well as trunk endpoint loss. The top and bottom images are from datasets SC-ISTI and SC-DUO, respectively.}
    \label{Fig:Ablation_fig}
    \vspace{-1em}
\end{figure*}

We conducted the ablation study on both the TISC-Net-m and TISC-Net-xP6 models to provide a more comprehensive validation. The evaluation scores on SC-ISTI and SC-DUO datasets are reported in Table \ref{tab:ablation}. On both datasets, our full models demonstrate superior performance in trunk identification compared to all ablated models, providing further evidence of the efficacy of FReLU, EMA and trunk endpoint loss. By comparing the object detection performance of -w/o-FReLU, -w/o-EMA, and the full model, we can observe that FReLU and EMA not only enhance the descriptive accuracy of the trunk but also effectively improve overall perception of the target area. The introduction of these key enhancements enables TISC-Net to achieve the coordination and collaborative optimization of object detection and curve prediction tasks within a unified framework. The visual examples of the ablation study are depicted in Fig. \ref{Fig:Ablation_fig}, while a detailed analysis is provided on a module-by-module basis as follows.

\begin{table*}[t]
\caption{Quantitative results of ablation study on SC-ISTI and SC-DUO datasets. We conducted ablation experiments with TISC-Net-m and TISC-Net-xP6 as full models, respectively, to verify the performance of ablation factors under models of different sizes. The best scores are in \textcolor{red}{red}. (Best viewed in color)\label{tab:ablation}}
\fontsize{9pt}{10pt}\selectfont
\renewcommand{\arraystretch}{1.5}
\centering
\setlength\tabcolsep{1.5pt}
\scalebox{0.9}{

\begin{tabular}{c|c|ccccclccccc}
\hline
                                   &                                      & \multicolumn{5}{c}{\textbf{SC-ISTI}}                                                                                                                     &  & \multicolumn{5}{c}{\textbf{SC-DUO}}                                                                                                                      \\ \cline{3-7} \cline{9-13} 
                                   &                                      & \multicolumn{2}{c}{\textbf{Object Detection}}               & \multicolumn{3}{c}{\textbf{Trunk Identification}}                                                  &  & \multicolumn{2}{c}{\textbf{Object Detection}}               & \multicolumn{3}{c}{\textbf{Trunk Identification}}                                                  \\
\multirow{-3}{*}{\textbf{Methods}} & \multirow{-3}{*}{\textbf{Ablations}} & \textbf{mAP50}$\uparrow$               & \textbf{mAP50-95}$\uparrow$            & \textbf{mAP50}$\uparrow$               & \textbf{mAP50-95}$\uparrow$            & \textbf{PCK}$\uparrow$                 &  & \textbf{mAP50}$\uparrow$               & \textbf{mAP50-95}$\uparrow$            & \textbf{mAP50}$\uparrow$               & \textbf{mAP50-95}$\uparrow$            & \textbf{PCK}$\uparrow$                 \\ \cline{1-7} \cline{9-13} 
                                   & -w/o-FReLU                           & 0.837                        & 0.528                        & 0.831                        & 0.775                        & 0.879                        &  & {\color[HTML]{FE0000} 0.701} & 0.271                        & 0.657                        & 0.531                        & 0.651                        \\
                                   & -w/o-EMA                             & 0.843                        & 0.574                        & 0.866                        & 0.809                        & 0.837                        &  & 0.668                        & 0.261                        & 0.677                        & 0.563                        & 0.680                        \\
                                   & -w/o-Endpointloss                    & 0.915                        & {\color[HTML]{FE0000} 0.593}                        & 0.847                        & 0.791                        & 0.856                        &  & 0.655                        & 0.276                        & 0.654                        & 0.562                        & 0.677                        \\
\multirow{-4}{*}{TISC-Net-m}        & \textbf{Full}                        & {\color[HTML]{FE0000} 0.929} & 0.591 & {\color[HTML]{FE0000} 0.890} & {\color[HTML]{FE0000} 0.828} & {\color[HTML]{FE0000} 0.920} &  & 0.695                        & {\color[HTML]{FE0000} 0.285} & {\color[HTML]{FE0000} 0.679} & {\color[HTML]{FE0000} 0.578} & {\color[HTML]{FE0000} 0.713} \\ \cline{1-7} \cline{9-13} 
                                   & -w/o-FReLU                           & 0.951                        & 0.638                        & 0.912                        & 0.857                        & 0.939                        &  & 0.753                        & 0.328                        & 0.725                        & 0.649                        & 0.750                        \\
                                   & -w/o-EMA                             & 0.957                        & 0.650                        & 0.901                        & 0.861                        & 0.928                        &  & 0.765                        & {\color[HTML]{FE0000} 0.337} & 0.707                        & 0.629                        & 0.761                        \\
                                   & -w/o-Endpointloss                    & {\color[HTML]{FE0000} 0.967} & {\color[HTML]{FE0000} 0.656}                        & 0.897                        & 0.866                        & 0.937                        &  & 0.767                        & 0.335                        & 0.733                        & 0.626                        & 0.773                        \\
\multirow{-4}{*}{TISC-Net-xP6}      & \textbf{Full}                        & 0.961                        & 0.653 & {\color[HTML]{FE0000} 0.914} & {\color[HTML]{FE0000} 0.871} & {\color[HTML]{FE0000} 0.941} &  & {\color[HTML]{FE0000} 0.772} & {\color[HTML]{333333} 0.331} & {\color[HTML]{FE0000} 0.738} & {\color[HTML]{FE0000} 0.658} & {\color[HTML]{FE0000} 0.776} \\ \hline
\end{tabular}
}
\end{table*} 

\subsubsection{Ablation Study on FReLU}
In our TISC-Net, we employ FReLU, which is designed as a 2D funnel-shaped activation, to enhance the local spatial dependency of deep features. For sea cucumber, which exhibits a relatively homogeneous appearance but is surrounded by an extremely complex and diverse background, incorporating FReLU enables us to establish more reliable spatial relationships, thereby facilitating improved object detection and trunk prediction. As reported in Table \ref{tab:comparison}, on the SC-ISTI dataset, the full TISC-Net-m demonstrates a remarkable performance improvement compared to its -w/o-FReLU counterpart. As shown in Fig. \ref{Fig:Ablation_fig} (a) and (b), the full model provides a more complete and well-fitted prediction of the trunk with better perception of local dependencies. For TISC-Net-xP6 and its -w/o-FReLU model, the improvement is less pronounced but FReLU still exhibits positive effectiveness. On the SC-DUO dataset, the superiority of the full models compared to their -w/o-FReLU counterparts mainly reflects on the trunk identification performance. 

\subsubsection{Ablation Study on EMA}
The EMA module, as a key component in the Neck part of our improved framework, contains 2D global average pooling and channel-wise attention. Through this module, our network can transfer attention across channels and establish cross-spatial dependency for simultaneous sea cucumber object detection and curve prediction. The ablation experiments verify its effectiveness in the two tasks, with a more pronounced improvement in performance specifically observed for the task of trunk identification.

Numerically, on the SC-DUO dataset, the full TISC-Net-xP6 model for trunk identification exhibits a obvious improvement in mAP50 and mAP50-95 scores compared to its -w/o-EMA counterpart, with an increase of over 3\% and 2\%, respectively. Additionally, the PCK score demonstrates an enhancement of 1.5\%. The effectiveness of EMA for trunk identification on the SC-ISTI dataset is also positive. The introduction of EMA has successfully rectified some failure cases in trunk prediction, as exemplified by Fig. \ref{Fig:Ablation_fig} (c).
\begin{figure*}[t]
\centering
	\includegraphics[width=0.95\linewidth]{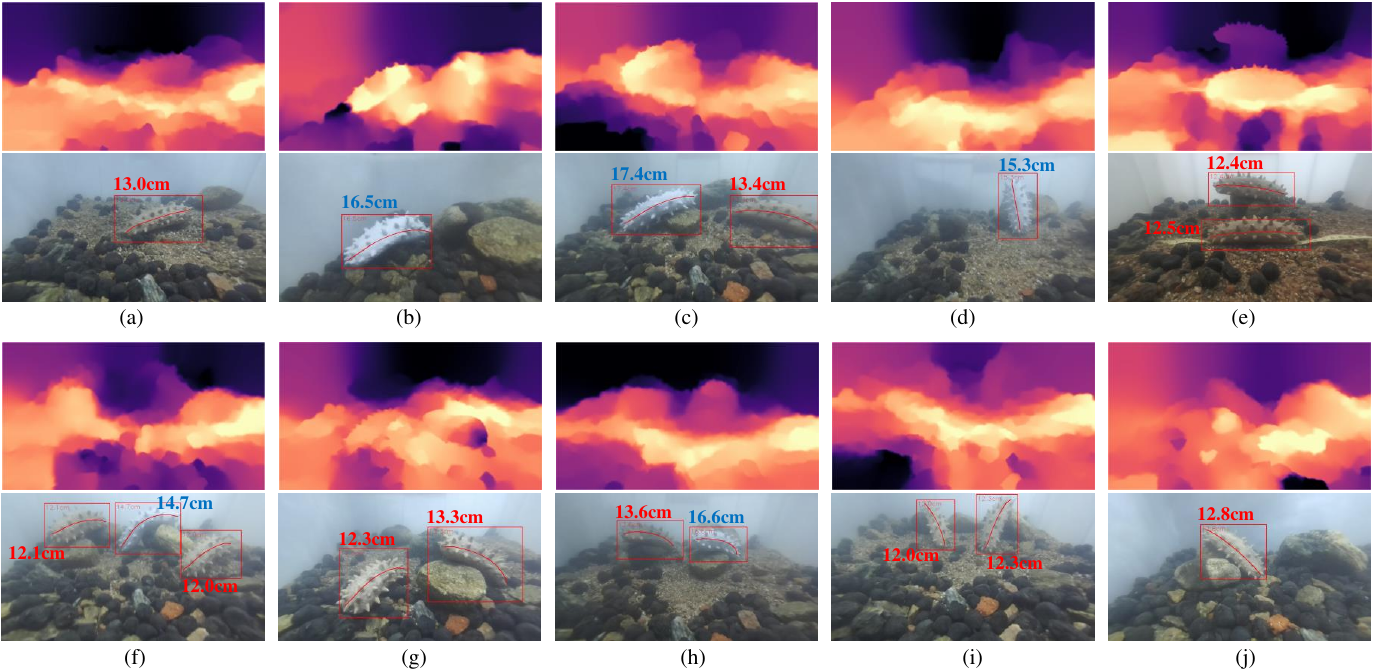}
    \caption{Visualized results of the measurements of sea cucumber lengths. Results of the pipeline using TISC-Net-m are presented. Each sample consists of the main view captured by the binocular camera and its corresponding depth map. For simulated sea cucumbers of the same length, the measurement results are indicated in the same font color. The true length of the \textcolor{red}{Red} marks is 12.5cm, and the true length of the \textcolor{blue}{Blue} marks is 16.0cm.}
    \label{Fig:length_depth}
    
\end{figure*}

\begin{figure*}[h]
  \centering
   \includegraphics[width=1\linewidth]{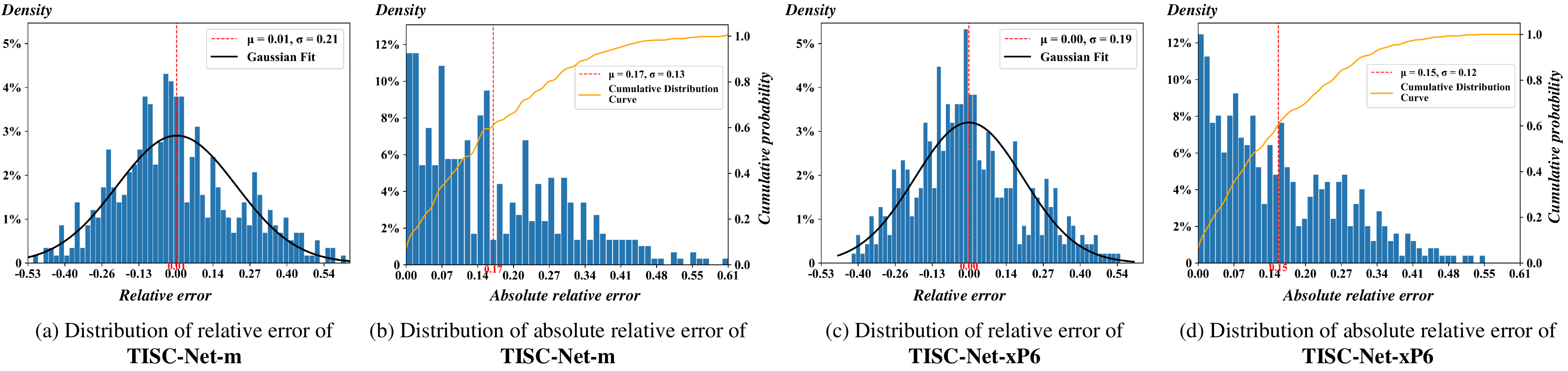}
   \caption{The distribution of length errors using the proposed in-situ trunk length measuring pipeline. The distributions of relative error in measurements are represented by (a) for TISC-Net-m and (c) for TISC-Net-xP6. We fit the distribution with Gaussian functions, whose mean errors and standard deviation are marked out. (b) and (d) are the distribution and cumulative distribution of absolute errors in measurements for TISC-Net-m and TISC-Net-xP6, respectively.}
   \label{fig:errorDistri}

\end{figure*}
\subsubsection{Ablation Study on Endpoint loss}\label{sec:ablation_endpointloss}
The trunk endpoint loss serves as an auxiliary term, aiming to reinforce the network's constraint on the accuracy of endpoint prediction.
The result of the two-stage method depicted in Fig.~\ref{Fig:comparison_of_methods} (a) serves as an illustrative example, highlighting how even a minor deviation in the endpoint can significantly compromise the accuracy of the curve.  If we eliminate the $\mathcal {L}_{epl}$ term, the curve will encounter greater challenges in accurately pinpointing (refer to Fig.~\ref{Fig:Ablation_fig} (a) and (d)), resulting in lower mAP50, mAp50-95, and PCK scores for trunk identification compared to our full method on both datasets. 

The earlier section of the article highlights the existence of a trade-off between object detection loss and curve prediction detection in our integrated framework. The evaluation of the full model and its -w/o-Enbpointloss models in Table~\ref{tab:ablation} demonstrates that the performance margin of trunk endpoint loss does not compromise the accuracy of object detection. This finding further illustrates the compatibility of this loss term with existing loss functions and learning strategies.

\subsection{Experiments on Length Measurement}
To quantitatively validate the effectiveness of the proposed visual measurement scheme for sea cucumber length, a simulated underwater environment was constructed in a water tank using simulated sea cucumbers, real sand, and rocks. Due to the stress response of real sea cucumbers, obtaining accurate body length data in underwater environments as a quantitative reference is challenging. Hence, in our experimental setup, we employ simulated models of sea cucumbers as a substitute. A binocular ZED camera, specifically calibrated for underwater use, was employed to record videos of sea cucumbers along with depth information. The video is stored in the SVO format, which is the original format of the ZED camera, without undergoing any post-processing. We extract $320$ valid images with depth information from the captured video. Out of the total $320$ images, a subset of $200$ RGB images is included in the training set for fine-tuning the TISC-Net model, while the remaining $120$ RGB images along with their corresponding depth maps are utilized for length measurement testing. 

In the testing phase, by utilizing the internal parameters of the binocular camera, we map the coordinates of the sea cucumber trunk curve identified by TISC-Net onto a three-dimensional coordinate system derived from the depth map. Subsequently, we quantitatively determine the length of the sea cucumber trunk through integration along its spatial curve as described in Eq. \eqref{integration_equation}. Several instances showcasing measurement results alongside their corresponding depth maps are illustrated in Fig. \ref{Fig:length_depth}. The true lengths of the simulated sea cucumbers marked in \textcolor{red}{Red} are $12.5$cm, and those marked in \textcolor{blue}{Blue} are $16.0$cm. 

To quantitatively assess the accuracy of the proposed in-situ sea cucumber length measurement pipeline, a statistical analysis is conducted on the measurement errors obtained from the aforementioned experiments. We assess the performance of the pipeline using TISC-Net-m and TISC-Net-xP6, respectively. For each of them, we present the distributions of relative error and absolute relative error in Fig. \ref{fig:errorDistri}. The relative error $e_{r}$ and absolute relative error $e_{ar}$ are defined as follows,
\begin{equation}
\label{eq:rela-error}
e_{r} = \frac{l_{mea} - l_{gt}}{l_{gt}}, \quad e_{ar} = \frac{\left | l_{mea} - l_{gt} \right | }{l_{gt}},
\end{equation}
where $l_{mea}$ and $l_{gt}$ are the measured length and its corresponding ground truth, respectively. According to the distribution of absolute relative errors, we further compute the cumulative probabilities for various levels of errors and depict them as curves, as demonstrated in Fig. \ref{fig:errorDistri} (b) and (d). We find that, the average absolute relative errors are $0.17$ and $0.15$, respectively, and more than $80\%$ of the test samples have absolute relative errors less than $0.3$. Moreover, the relative error of TISC-Net-m based and TISC-Net-xP6 based pipelines, as observed from Fig. \ref{fig:errorDistri} (a) and (c), both exhibit normal distributions centered around zero mean. This implies that the proposed pipeline can achieve highly robust length measurement results by averaging multiple measurements obtained from various perspectives in real-world application scenarios. 

As shown in the tail of the error distribution, we also notice that a small number of results exhibit significant measurement errors. The depth information collected by the binocular camera, often exhibits distortions such as sudden changes and gaps in depth, as shown in Fig.~\ref{Fig:length_depth} (b), (c), and (i), which significantly contribute to the observed large measurement error. Besides, the depth information of the pixel where the curve is located lacks stability, and the curve may extend beyond the effective area of the sea cucumber's main body. These factors also contribute to fluctuations in measurement accuracy, as evidenced by the distribution tails. Moreover, part of the sea cucumber trunk is in the blind spot at certain viewing angles, which also affects the accuracy of measurements (see Fig.~\ref{Fig:length_depth} (d)). In fact, as we discussed earlier, these issues can be easily resolved by taking multiple measurements and using their mean value to obtain a reliable length of sea cucumbers.

\section{Discussion and Future works}\label{sec:discussion}
As illustrated in Fig. \ref{Fig:datasets_example}, the geometric properties of the sea cucumber trunks can be effectively described by B\'{e}zier curves. The proposed method demonstrates higher accuracy in the task of trunk curve characterization, as indicated by the results of quantitative evaluation, compared to the comparison schemes. While, as a common challenge in curve modeling, it is difficult for the curvature coefficients to generalize when there is a highly biased distribution of curvatures \citep{Rethinking_Bezier_curve}. An example can be seen in Fig.~\ref{Fig:datasets_example} (f) of the SC-ISTI dataset. The problem can be partially mitigated by increasing the proportion of curve loss. However, this adjustment inevitably entails a drawback, leading to target omission. This trade-off is a common challenge in all one-stage joint framework methods. In future research, it would be interesting to discover improved constraints in order to achieve better generalization for large bends. Additionally, exploring how to strike a balance among multiple tasks is also a highly promising topic.

As a one-stage approach for sea cucumber trunk identification, it demonstrates high efficiency (FPS $>$ 35 for TISC-Net-xP6 and FPS $>$ 65 for TISC-Net-m) and deployability on edge computing devices, as depicted in Table \ref{tab:Parameters}. The high efficiency of our system allows us to utilize a diverse range of options in order to enhance the precision of in-situ measurements when deploying it on an ROV. For instance, employing multi-viewing and multiple measurements fusion can further effectively mitigate measurement errors, which may caused by the inevitable degradation of depth estimation in underwater scenes. The exploration of the aforementioned strategies holds great promise for future investigation.

\section{Conclusion}
The paper introduces TISC-Net, a new framework based on B\'{e}zier curve modeling for accurately identifying sea cucumber trunks. Additionally, a length measurement pipeline is designed to facilitate in-situ resource surveying of sea cucumbers. To the best of our knowledge, this study presents the first sea cucumber trunk identification scheme with high efficiency and introduces B\'{e}zier curves for skeleton recognition. We thoroughly considered the scene characteristics of sea cucumber natural habitat images and innovatively integrate B\'{e}zier curve modeling with deep detectors to achieve end-to-end trunk identification. By further introducing FReLU activation and EMA module to improve curve feature perception and learning ability, along with the endpoint loss function to strengthen the constraint on the endpoints, we achieved robust in-situ trunk curve modeling and identification. The effectiveness of our method is demonstrated through comprehensive comparison and ablation studies on two new benchmarks, namely SC-ISTI and SC-DUO. In addition, we build a simulated underwater habitat scene for sea cucumbers and validated the accuracy of in-situ length measurement through the proposed pipeline. The proposed unified framework demonstrates high efficiency and deployability, indicating its potential for wide practical application. Furthermore, future enhancements can be achieved by incorporating the fusion of multi-view or multiple measurements.

\section*{Acknowledgment}
The research has been supported by the National Natural Science Foundation of China under Grant 62371431 and 61906177, in part by the Marine Industry Key Technology Research and Industrialization Demonstration Project of Qingdao under Grant 23-1-3-hygg-20-hy, in part by the Significant Applied Technology Innovation Projects for Agriculture of Shandong Province under Grant SD2019NJ020, and in part by the Fundamental Research Funds for the Central Universities under Grants 202262004.




\bibliographystyle{elsarticle-harv}

\bibliography{reference}


\end{document}